\documentclass[lettersize,journal]{IEEEtran}
\usepackage{amsmath,amsfonts}
\usepackage{algorithmic}
\usepackage{algorithm}
\usepackage{array}
\usepackage[caption=false,font=normalsize,labelfont=sf,textfont=sf]{subfig}
\usepackage{textcomp}
\usepackage{stfloats}
\usepackage{url}
\usepackage{verbatim}
\usepackage{graphicx}
\usepackage{cite}
\usepackage{bm}
\usepackage{bbding}
\usepackage{algorithm}
\usepackage{algorithmic}
\begin{document}

\title{Reward-Guided Semantic Evolution for Test-time Adaptive Object Detection}

\author{Lihua Zhou,  
Mao Ye,~\IEEEmembership{Senior Member,~IEEE,}
Xiatian Zhu,
Nianxin Li,
Changyi Ma,
Shuaifeng Li,\\
Yitong Qin,
Hongbin Liu,
Jiebo Luo,~\IEEEmembership{Fellow,~IEEE,}
Zhen Lei$^*$, ~\IEEEmembership{Fellow,~IEEE} 
\thanks{
\IEEEcompsocthanksitem Lihua Zhou and Hongbin Liu are with the Centre for Artificial Intelligence and Robotics, Hong Kong Institute of Science and Innovation, Chinese Academy of Sciences, Hong Kong, China.
Email: lihuazhou120@gmail.com, liuhongbin@ia.ac.cn
\IEEEcompsocthanksitem Mao Ye, Nianxin Li, Shuaifeng Li and Yitong Qin are with School of Computer Science and Engineering, University of Electronic Science and Technology of China, Chengdu 611731, China. E-mail: maoye@uestc.edu.cn, hotwindlsf@gmail.com, linianxin1220@gmail.com
\IEEEcompsocthanksitem Xiatian Zhu is with Surrey Institute for People-Centred Artificial Intelligence, CVSSP, University of Surrey, Guildford, UK. E-mail: xiatian.zhu@surrey.ac.uk
\IEEEcompsocthanksitem {Changyi Ma} is with the School of Artificial Intelligence, 
Jilin University, China. Email: changyima@jlu.edu.cn
\IEEEcompsocthanksitem {Jiebo Luo} is with the University of Rochester and performed this work while on a sabbatical leave at the Hong Kong Institute of Science and Innovation.
\IEEEcompsocthanksitem Zhen Lei is with the School of Artificial Intelligence, University of Chinese Academy of Sciences (UCAS), Beijing 100049, China; the Centre for Artificial Intelligence and Robotics, Hong Kong Institute of
Science and Innovation, Chinese Academy of Sciences, Hong Kong, China.
Email: zhen.lei@ia.ac.cn
\IEEEcompsocthanksitem * corresponding author.
}
}

\markboth{Journal of \LaTeX\ Class Files,~Vol.~14, No.~8, August~2021}%
{Shell \MakeLowercase{\textit{et al.}}: A Sample Article Using IEEEtran.cls for IEEE Journals}


\maketitle

\begin{abstract}
Open-vocabulary object detection with vision-language models (VLMs) such as Grounding DINO suffers from performance degradation under test-time distribution shifts, primarily due to semantic misalignment between text embeddings and shifted visual embeddings of region proposals. While recent test-time adaptive object detection methods for VLM-based either rely on costly backpropagation or bypass semantic misalignment via external memory, none directly and efficiently align text and vision in a training-free manner. To address this, we propose Reward-Guided Semantic Evolution (RGSE), a training-free framework that directly refines the text embeddings at test time. Inspired by evolutionary search, RGSE treats text embedding adaptation as a semantic search process: it perturbs text embeddings as candidate variants, evaluates them via cosine similarity with current and historical high-confidence visual proposals as a reward signal, and fuses them into a refined embedding through reward-weighted averaging. Without any backpropagation, RGSE achieves state-of-the-art performance across multiple detection benchmarks while adding minimal computational overhead. Our code will be open source upon publication. 
\end{abstract}


\section{Introduction}\label{sec:intro}

Vision-language models (VLMs), such as CLIP \cite{radford2021learning} and Grounding DINO \cite{liu2024grounding}, have enabled powerful open-vocabulary perception by aligning image and text representations in a shared semantic space. This alignment facilitates zero-shot generalization to novel categories without task-specific training \cite{li2025generalizing}. However, their performance often degrades significantly under distribution shifts between training and test data, which are common in real-world scenarios \cite{pan2009survey,tian2026dual,shao2025consistent,wang2025deep,lu2025adaptive}. To address this, test-time adaptation (TTA) \cite{liang2025comprehensive,liu2025towards,wu2025a3} has emerged as an effective strategy that adapts model predictions using only unlabeled test samples.

\begin{figure}[t]
    \centering
    \includegraphics[width=0.38\textwidth]{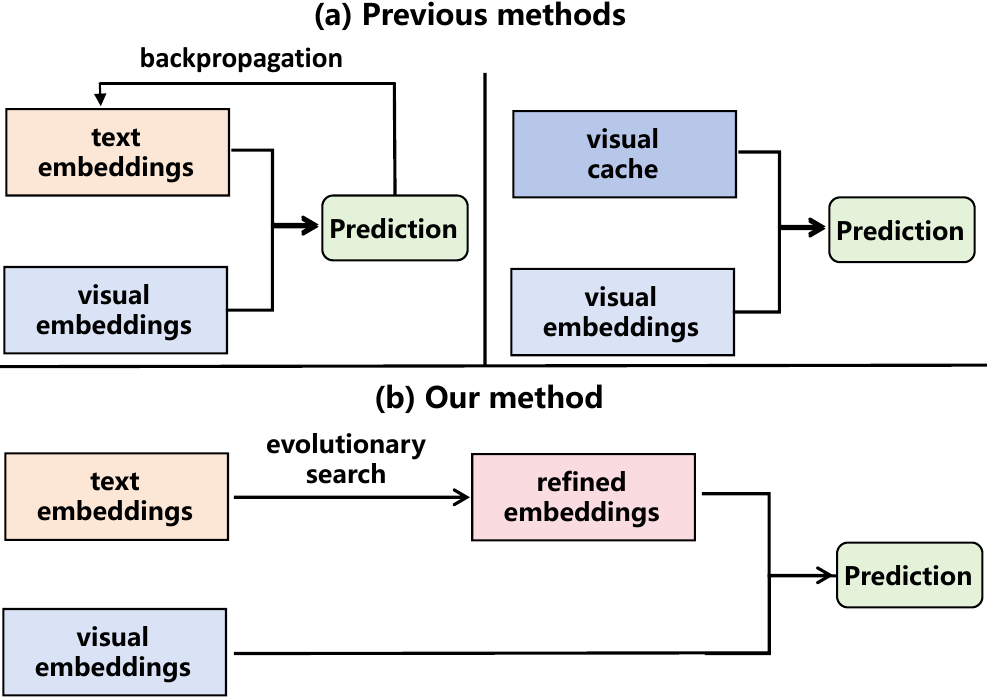}
    \caption{Comparison between previous methods and RGSE.  (a) Previous methods either rely on computationally expensive backpropagation to update text embeddings or bypass semantic misalignment by maintaining a separate visual cache. (b) RGSE treats text embedding adaptation as an evolutionary search, which enables direct and efficient correction of semantic misalignment.}
    \label{fig:intro}
    \vspace{-0.5cm}
\end{figure}

While TTA has been extensively studied for VLM-based classification \cite{shu2022test,feng2023diverse,karmanov2024efficient,sheng2025r,sharifdeen2025tpt,zhou2025bayesian,zhang2024dual,zhang2025cola}, its application to object detection remains limited, with little effort exploring TTA for vision-language object detectors \cite{belal2025vlod,zhou2025bayesianarxiv,zhang2024historical,gaotest}, which can be broadly categorized into two groups based on whether they require training during inference.
The first group consists of approaches that rely on backpropagation-based optimization at inference time.
For example, VLOD-TTA \cite{belal2025vlod} uses IoU-weighted entropy minimization and image-conditioned prompt selection; HisTPT \cite{zhang2024historical} applies prompt tuning with historical memory banks to mitigate catastrophic forgetting; and MPMT \cite{gaotest} adopts a multi-modal prompt-based Mean-Teacher framework with instance-level dynamic memory.
The second group comprises training-free methods, such as BCA+ \cite{zhou2025bayesianarxiv}, which proposes a Bayesian framework that maintains a cache, then fuses cache-based predictions with the VLM outputs.

Despite these advances, existing methods \cite{belal2025vlod,zhou2025bayesianarxiv,zhang2024historical} fail to directly and efficiently correct the semantic misalignment between text embeddings and shifted visual embeddings during test time. As shown in Fig. \ref{fig:intro}(a), VLOD-TTA \cite{belal2025vlod}, HisTPT \cite{zhang2024historical} and MPMT \cite{gaotest} attempt to refine text embeddings, but they rely on backpropagation during inference, which introduces significant computational overhead and undermines the real-time requirement of TTA. In contrast, 
BCA+ \cite{zhou2025bayesianarxiv} avoids gradient computation by maintaining a dynamic cache of historical visual embeddings and adaptive priors. However, it bypasses the misalignment issue rather than correcting it, as the original text embeddings remain unchanged throughout adaptation. 
Consequently, none of the current methods directly and efficiently refine the text embeddings to better align with the test-domain visual context.

To address this gap, we propose Reward-Guided Semantic Evolution (RGSE), a training-free method that directly refines the text embeddings of vision-language object detectors at test time as shown in Fig. \ref{fig:intro}(b). Inspired by the principle of reward-based optimization of evolutionary search \cite{hansen2016cma,jaderberg2017population}, RGSE treats text embedding adaptation as a semantic search process: for each test image, we first obtain the text embeddings and region proposal visual embeddings from the vision-language object detector, such as Grounding DINO \cite{liu2024grounding}. We then generate multiple perturbed versions of the text embeddings, which serve as candidate variants in the semantic search process. Each perturbed text embedding is evaluated by its alignment with the current visual context, which includes both the proposals from the current image and a memory of historically high-confidence proposals. This alignment, measured via cosine similarity, serves as an instantaneous reward that reflects how well the perturbed text embedding matches the visual semantics of the test domain. Finally, a refined text embedding is obtained as a reward-weighted combination of perturbed candidates. This process requires no backpropagation, adds minimal computation, and directly corrects the semantic misalignment.

Our main contributions are summarized as follows:
(1) We propose Reward-Guided Semantic Evolution (RGSE), a training-free framework that directly refines the original text embeddings of VLM-based detectors at test time without backpropagation; 
(2) We formulate text embedding adaptation as a semantic search process, inspired by evolutionary search via perturbation and reward-guided reweighting, and 
(3) RGSE achieves state-of-the-art performance on multiple detection benchmarks, with ablation studies further demonstrating its efficiency and robustness.

\section{Related Work}\label{sec:related_work}

\noindent\textbf{Vision-language models.}
Early deep learning approaches typically require training a separate neural network for each vision task, which is a labor-intensive process that lacks generalization across domains \cite{he2016deep,ren2016faster}. This paradigm has been fundamentally challenged by the rise of Vision-Language Models (VLMs), which learn joint multi-modal representations from massive web-scale image-text pairs and enable zero-shot inference across diverse tasks with a single model \cite{radford2021learning,jia2021scaling}. Pioneering work like CLIP \cite{radford2021learning} aligns global image and text embeddings via contrastive learning, allowing zero-shot image classification by matching visual embeddings to textual category names. Subsequent models such as ALIGN~\cite{jia2021scaling}, FILIP~\cite{yaofilip}, PaLI~\cite{chenpali}, LiT~\cite{zhai2022lit}, and ZeroVL~\cite{cui2022contrastive} further scale up training data or refine cross-modal alignment, establishing VLMs as a powerful foundation for open-vocabulary classification. 

The success of VLMs in classification has inspired their extension to open-vocabulary object detection. Early efforts adopt knowledge distillation: ViLD \cite{gu2022open} transfers CLIP’s image-level knowledge to a two-stage detector, while DetCLIP \cite{yao2022detclip} leverages caption data to generate pseudo bounding-box labels. More recent works pursue end-to-end grounding, treating detection as phrase-region alignment. 
OV-DETR \cite{zareian2021open} leverages image and text embeddings from CLIP as learnable queries in the DETR framework \cite{carion2020end}, enabling it to generate category-specific bounding boxes directly from language inputs.
GLIP \cite{li2022grounded} unifies detection and phrase grounding via a shared contrastive loss, achieving strong zero-shot transfer. Grounding DINO \cite{liu2024grounding} further advances this line by integrating language-guided query selection and cross-modal feature enhancement, setting a new state of the art in open-vocabulary detection without task-specific fine-tuning.
YOLO-World \cite{cheng2024yolo} introduces reparameterizable vision–language fusion for real-time open-vocabulary detection, enabling deployment in latency-sensitive scenarios.

\noindent\textbf{Test-time adaptive object detection.} 
Early test-time adaptive object detection methods focus on closed-set detectors such as Faster R-CNN \cite{ren2016faster}. Pioneering works such as TENT \cite{wang2021tent} adapted batch normalization statistics by minimizing predictive entropy. Subsequent approaches like STFAR \cite{chen2023stfar} introduced self-training paradigms with feature distribution regularization to mitigate noisy pseudo-labels in dynamic environments. 
W3TTAOD \cite{yoo2024and} proposed an architecture-agnostic method using lightweight adapters and stability-aware objectives for efficient updates.
MemCLR \cite{vs2023towards} improves target-specific representations via contrastive learning in dynamic environments.
MLFA \cite{liu2024mlfa} adapts models online by matching both the global-level and category cluster-level distributions of informative features across different domains.
CTTA-OD \cite{wang2025efficient} introduces a sensitivity-guided channel pruning strategy that quantifies each channel based on its sensitivity to domain discrepancies. 
IoU Filter \cite{ruan2024fully} proposes a pseudo-label filtering by matching detections across consecutive self-training iterations and removing duplicate detections with conflicting class predictions.
These early methods demonstrated effectiveness but were often coupled with specific architectures like Faster R-CNN \cite{ren2016faster} and relied on computationally intensive backpropagation, limiting real-time applicability.

Recently, test-time adaptive object detection methods have been extended to vision-language models such as Grounding DINO \cite{liu2024grounding}.
VLOD-TTA \cite{belal2025vlod} improves robustness via IoU-weighted entropy minimization and image-conditioned prompt selection. 
HisTPT \cite{zhang2024historical} mitigates catastrophic forgetting in prompt tuning by maintaining multiple knowledge banks of historical test samples.
MPMT \cite{gaotest} adopts a multi-modal prompt-based Mean-Teacher framework with dynamic instance memory to jointly adapt visual and textual representations.
However, these methods relies on gradient-based optimization. In contrast, BCA+ \cite{zhou2025bayesianarxiv} proposes a training-free Bayesian framework that fuses the original VLM prediction with a cache-based prediction derived from historical class embeddings, spatial scales, and adaptive priors. While BCA+ avoids backpropagation, it does not refine the original text embeddings and instead corrects predictions through post-hoc fusion.

\noindent\textbf{Evolutionary search.}
Evolutionary search is a class of gradient-free optimization methods inspired by natural selection, where candidate solutions are perturbed, evaluated via a fitness function, and recombined to generate improved offspring over iterations. Classical approaches like CMA-ES \cite{hansen2016cma} maintain a distribution over promising solutions and update it using performance feedback, without relying on gradients. In computer vision, evolutionary strategies have been applied to hyperparameter tuning \cite{jaderberg2017population}, neural architecture search \cite{real2017large,real2019regularized}, and feature selection \cite{xue2015survey}.

Our work draws inspiration from this paradigm: instead of optimizing text embedding via prompt tuning, we treat text embedding refinement as a one-step evolutionary process. We generate perturbed text embeddings as candidate solutions, evaluate them using alignment with visual proposals as a fitness (reward) signal, and fuse them into a refined embedding via reward-weighted recombination. This allows directly adapting the semantic anchor of VLM to the test-domain visual context without gradient computation.

\section{Method}
\subsection{Preliminary}
\noindent \textbf{Problem setting.} In this work, we address the problem of test-time adaptive object detection for vision-language object detectors under distribution shift. Specifically, given a pre-trained VLM such as Grounding DINO \cite{liu2024grounding} and a stream of unlabeled test images \(\{x_i\}_{i=1}^I\) drawn from a domain that differs from the pretraining distribution, our goal is to adapt the model’s predictions online, without backpropagation, and without access to source data.

\noindent\textbf{Grounding DINO.} Grounding DINO is an open-vocabulary object detection model based on vision-language alignment.
In Grounding DINO, for an input image \(x_i\) and a set of \(K\) category names \(\mathcal{C} = \{c_k\}_{k=1}^K\) (e.g., “car”, “person”), the model first encodes the prompts into vanilla text embeddings using a text encoder (e.g., BERT \cite{devlin2019bert}), and extracts vanilla visual embeddings from the image using a visual encoder (e.g., Swin Transformer \cite{liu2021swin}). These are then jointly enhanced through a cross-attention-based feature enhancer, yielding text embeddings \(\bm{T} = \{ \bm{t}_k \}_{k=1}^K \in \mathbb{R}^{K \times d}\) and visual embeddings \(\bm{V} \in \mathbb{R}^{N_I \times d}\), where $d$ is the embedding dimension and $N_I$ represents the number of image tokens (typically over 10,000). From \(\bm{V}\), the model further selects 900 region proposals via language-guided query selection, and refines them through a cross-modality decoder \cite{zhangdino} to produce proposal visual embeddings \(\{ \bm{v}_m \}_{m=1}^{900} \) and bounding boxes \(\{ \bm{b}_m \}_{m=1}^{900}\). The initial classification score for the \(m\)-th proposal is computed as:  

\begin{equation}
    \bm{s}_m = \sigma( \mathrm{cos}(\bm{v}_m,\bm{T})),
    \label{eq:gdino_score}
\end{equation}
where $\mathrm{cos}(\cdot, \cdot)$ denotes the cosine similarity function and \(\sigma(\cdot)\) is the sigmoid function. To facilitate subsequent processing, we then convert these raw scores into a valid probability distribution format using the softmax function, yielding the initial predictions \(\bm{p}_m^{\text{init}}\):
\begin{equation}
    \begin{aligned}
        \bm{p}_m^{{init}} = \mathrm{softmax}(\bm{s}_m).
        \label{eq:gdino}
    \end{aligned}
\end{equation}
To obtain reliable detections, a confidence threshold is applied to filter out low-scoring proposals.
This yields a set of \(M\) high-confidence proposals, with visual embeddings \(\{ \bm{v}_m \}_{m=1}^{M} \), bounding boxes \(\{ \bm{b}_m \}_{m=1}^{M} \), initial predictions \(\{ \bm{p}_m^{\text{init}} \}_{m=1}^{M} \), 
and pseudo-labels \( \{{l}_m\}_{m=1}^M \), 
where \({l}_m = \arg\max_{k} \bm{p}_m^{\text{init}}[k]\). 

\begin{figure*}[t]
    \centering
    \includegraphics[width=0.8\textwidth]{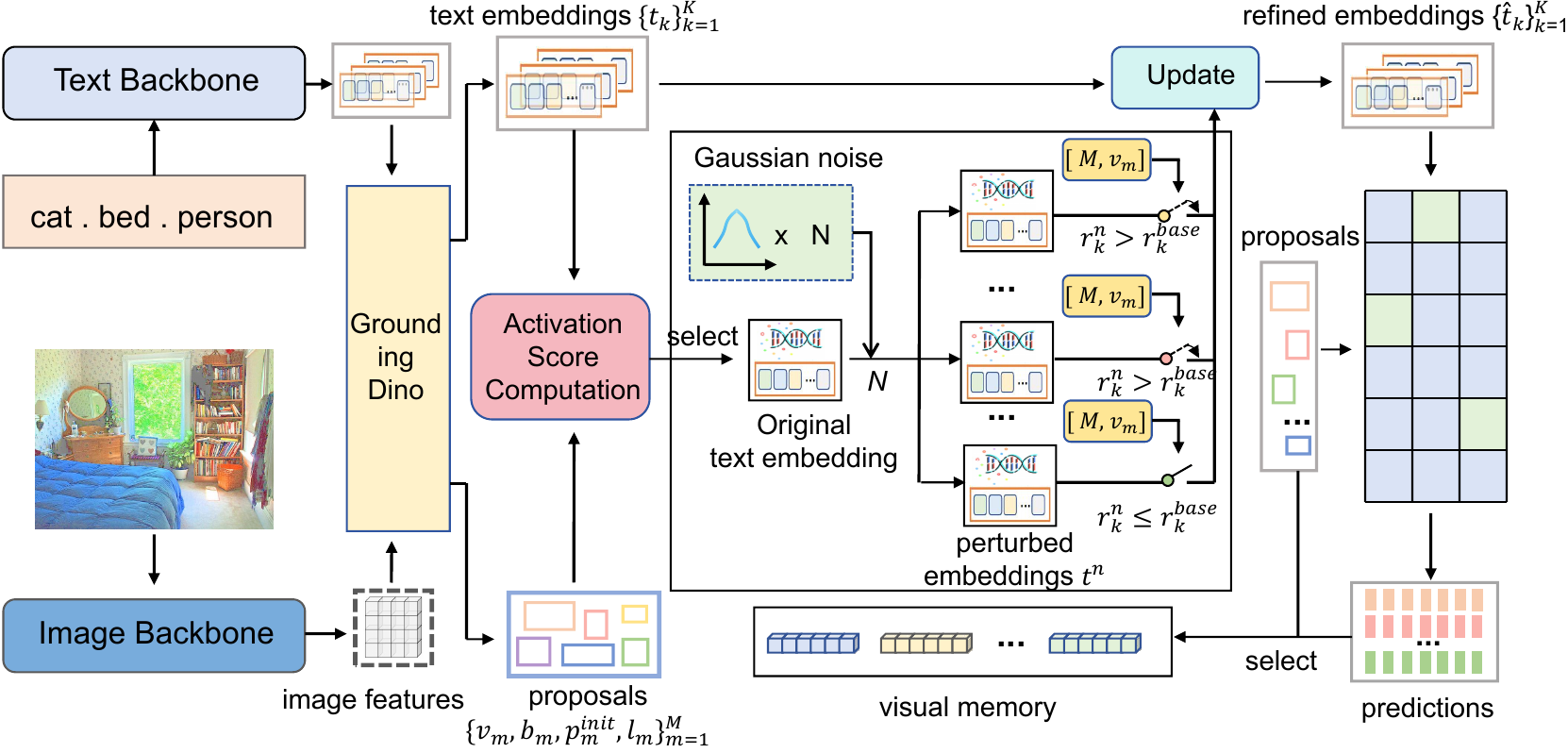}
    \caption{Overview of Reward-Guided Semantic Evolution (RGSE). Given a test image, we first obtain the initial outputs from Grounding DINO: region proposals $\{\bm{v}_m, \bm{b}_m, \bm{p}_m^{\text{init}}, l_m\}_{m=1}^M$, and text embeddings $\{\bm{t}_k\}_{k=1}^K$. To address semantic misalignment under domain shift, RGSE refines the original text embeddings through four forward-pass stages: (1) Perturbation: Gaussian noise is added to active text embeddings; (2) Reward evaluation: each perturbed embedding is scored based on cosine similarity with current and historical high-confidence visual proposals; (3) Refinement: only candidates outperforming the parent are fused via reward-weighted averaging; (4) Memory update: high-confidence proposals are added to a class-specific cache for future adaptation.}
    \label{fig:overview}
\end{figure*}

\subsection{Reward-Guided Semantic Evolution}
Given a test image, we start from the outputs of Grounding DINO: \(M\) region proposals with visual embeddings \(\{ \bm{v}_m \}_{m=1}^{M}\), bounding boxes \(\{ \bm{b}_m \}_{m=1}^{M}\), initial predictions \(\{ \bm{p}_m^{\text{init}} \}_{m=1}^{M}\), pseudo-labels \(\{ l_m \}_{m=1}^{M}\), and \(K\) text embeddings \(\{ \bm{t}_k \}_{k=1}^{K}\). However, under domain shift, the text embeddings may misalign with the shifted visual embeddings, leading to degraded detection performance.
To address this, we propose Reward-Guided Semantic Evolution (RGSE), a test-time adaptation framework that refines the text embeddings inspired by evolutionary search through a four-stage process as shown in Fig. \ref{fig:overview}:  
(1) Perturbation: We generate multiple candidate text embeddings by adding Gaussian noise to selected \(\bm{t}_k\);  
(2) Reward evaluation: For each perturbed embedding, we compute a reward as the cosine similarity between it and the visual embeddings of both current proposals and historically high-confidence proposals stored in a lightweight memory;  
(3) Refinement: We fuse all perturbed embeddings into a refined version via reward-weighted averaging;  
(4) Memory update: High-confidence proposals from the current image are added to the memory with a fixed-size queue to control growth.  
This process is fully forward-pass based, requiring no backpropagation.

\noindent\textbf{Perturbation.} 
Under domain shift, the text embeddings may misalign with the visual embeddings, leading to degraded initial predictions \(\{ \bm{p}_m^{\text{init}} \}_{m=1}^{M}\) and reduced detection performance. To address this, we adopt principles from evolutionary search \cite{hansen2016cma,jaderberg2017population}: rather than assuming the embedding is optimal, we treat it as a “parent” and generate “offspring” through local perturbation, allowing the semantic representation to evolve toward better alignment with the current visual context.

To identify which categories are likely present in the current image, we compute an activation score $s_k$ for each class using the initial predictions \(\{ \bm{p}_m^{\text{init}} \}_{m=1}^{M}\). The activation score is defined as the maximum confidence over all proposals:
\begin{equation}
    \begin{aligned}
    s_k = \max_{1 \leq m \leq M} \, \bm{p}_m^{\text{init}}[k]
    \label{eq:rbase}
    \end{aligned}
\end{equation}
We then select the set of active categories as \(\mathcal{K}_{\text{active}} = \{ k \mid s_k > \tau_{base} \}\), where \(\tau_{base}\) is a threshold for selecting likely categories, and perturb only their text embeddings.

For inactive classes (\(k \notin \mathcal{K}_{\text{active}}\)), we keep the text embedding unchanged.
For each active class \(k \in \mathcal{K}_{\text{active}}\), we generate \(N\) perturbed text embeddings by adding independent Gaussian noise \(\boldsymbol{\epsilon}_k^{n} \sim \mathcal{N}(0, \sigma^2 \mathbf{I})\):
\begin{equation}
    \begin{aligned}
    \tilde{\bm{t}}_k^{n} = \bm{t}_k + \boldsymbol{\epsilon}_k^{n}, \quad n = 1, \dots, N.
    \label{eq:Perturbation}
    \end{aligned}
\end{equation}
This yields \(N\) candidate text embedding sets \(\{ \tilde{\bm{t}}_k^{n} \}_{n=1}^N\), each representing a semantic variant to be evaluated in the next stage.
In our experiments, we typically set \(N\) to a relatively large value, such as {1000}, to expand the search space and increase the likelihood of identifying effective semantic variants. Thanks to efficient matrix operations, this does not significantly increase computational overhead. Moreover, the additional memory usage for storing these candidates is negligible compared to the model weights. We provide a detailed analysis of these aspects in the {Sec. \ref{sec:ablation}}.

\noindent\textbf{Reward evaluation.} For each active class \(k \in \mathcal{K}_{\text{active}}\) and each perturbed embedding \(\tilde{\bm{t}}_k^{n}\), we compute a reward that measures its similarity with the current visual context. Let \(\mathcal{P}_k = \{ m \mid l_m = k \}\) be the set of proposals pseudo-labeled as class \(k\). The visual context comprises:
(i) the visual embeddings for class \(k\), i.e., \(\{ \bm{v}_m \}_{m \in \mathcal{P}_k}\);  
(ii) a memory bank \(\mathcal{M}_k = \{ \bm{v}_m^{{hist}_k} \}_{m=1}^{H_k}\) storing visual embeddings of historically high-confidence proposals for class \(k\), where \(H_k\) is the size of the memory bank for class \(k\).
The reward for \(\tilde{\bm{t}}_k^{n}\) is defined as a weighted sum of cosine similarities:
\begin{equation}
    r_k^{n} =  \frac{1}{|\mathcal{P}_k|} \sum_{m \in \mathcal{P}_k}\mathrm{cos}\big( \tilde{\bm{t}}_k^{n}, \bm{v}_m \big) +   \frac{\alpha}{H_k} \sum_{m=1}^{H_k} \mathrm{cos}\big( \tilde{\bm{t}}_k^{n}, \bm{v}_m^{{hist}_k} \big),
    \label{eq:reward}
\end{equation}
where \(\alpha\) balances contributions of current and historical information. This reward reflects how well the perturbed text embedding aligns with both the current visual embeddings and the accumulated knowledge from past test samples.

\noindent\textbf{Refinement.}
For each active class \(k \in \mathcal{K}_{\text{active}}\), we first compute the base reward \(r^{\text{base}}_k\) for the original text embedding \(\bm{t}_k\) using Eq.~\eqref{eq:reward}. This serves as a performance baseline for the ``parent'' embedding. 
To mimic natural selection, we discard any perturbed embedding whose reward is lower than the base reward, i.e., \(r_k^{n} \leq r^{\text{base}}_k\). This ensures that only ``offspring'' that outperform the ``parent'' in aligning with the current visual context are retained for fusion.
Let \(\mathcal{N}_k^{\text{survive}} = \{ n \mid r_k^{n} > r^{\text{base}}_k \}\) denote the set of surviving candidates. The refined text embedding for class \(k\) is then computed as a reward-weighted average over the survivors:
\begin{equation}
\begin{aligned}
    &\tilde{\bm{t}}_k = w_k^{\text{base}} \cdot \bm{t}_k+\sum_{n \in \mathcal{N}_k^{\text{survive}}} w_k^{n} \cdot \tilde{\bm{t}}_k^{n}, \\
    \label{eq:refinement}
\end{aligned}
\end{equation}
where $w_k^{\text{base}} = \frac{\exp(r_k^{\text{base}})}{ \exp(r_k^{\text{base}}) + \sum_{n' \in \mathcal{N}_k^{\text{survive}}} \exp(r_k^{n'})}$ and $w_k^{n} = \frac{\exp(r_k^{n})}{ \exp(r_k^{\text{base}}) + \sum_{n' \in \mathcal{N}_k^{\text{survive}}} \exp(r_k^{n'})}$.
We maintain a global text embedding bank \(\bar{\bm{T}} = \{ \bar{\bm{t}}_k \}_{k=1}^K\) and a counter \(\{ n_k \}_{k=1}^K\), both initialized to zero: \(\bar{\bm{t}}_k = \mathbf{0}\) and \(n_k = 0\) for all \(k\). For each active class \(k \in \mathcal{K}_{\text{active}}\), we update its global embedding and counter as a running average:
\begin{equation}
\begin{aligned}
\bar{\bm{t}}_k = \frac{n_k \cdot \bar{\bm{t}}_k + \tilde{\bm{t}}_k}{n_k + 1}, \quad n_k = n_k + 1.
    \label{eq:update}
\end{aligned}
\end{equation}
For inactive classes (\(k \notin \mathcal{K}_{\text{active}}\)), both \(\bar{\bm{t}}_k\) and \(n_k\) remain unchanged.
Finally, we construct a prediction embedding set \(\hat{\bm{T}} = \{ \hat{\bm{t}}_k \}_{k=1}^K\), where \(\hat{\bm{t}}_k = \bar{\bm{t}}_k\) if \(n_k > 0\) (indicating the global embedding has been updated) and \(\hat{\bm{t}}_k = \bm{t}_k\) if \(n_k = 0\) (falling back to the original text embedding). This \(\hat{\bm{T}}\) is then used to recompute the final classification scores. For the \(m\)-th proposal with visual embedding \(\bm{v}_m\), the refined probability is:
\begin{equation}
    \begin{aligned}
        \bm{p}_m = \mathrm{softmax}\big( \sigma( \mathrm{cos}(\bm{v}_m,\hat{\bm{T}})) \big).
        \label{eq:refineprediction}
    \end{aligned}
\end{equation}

\begin{algorithm}[t]
\caption{Reward-Guided Semantic Evolution}
\label{alg:method}
\begin{algorithmic}[1]  
\STATE \textbf{Input:} Pre-trained Grounding DINO, test images $\{x_i\}_{i=1}^I$, hyperparameters $\tau_{{base}}$, $N$, $\sigma$, $\alpha$, $M_{max}$
\STATE \textbf{Initialize:} Global text embedding bank $\bar{\bm{T}} = \mathbf{0}$, counter $n_k = 0$, memory banks $\mathcal{M}_k = \emptyset$ for all $k$

\FOR{each test image $x_i$}
    \STATE Obtain $K$ text embeddings \(\{ \bm{t}_k \}_{k=1}^{K}\) and $M$ high confidence proposals $\{\bm{v}_m, \bm{b}_m, \bm{p}_m^{\text{init}}, l_m\}_{m=1}^M$
    
    \STATE Compute activation score $s_k$ and select the set of active categories $\mathcal{K}_{\text{active}}$
    
    \FOR{$k \in \mathcal{K}_{\text{active}}$}
        \STATE Generate $N$ perturbed embeddings \(\{ \tilde{\bm{t}}_k^{n} \}_{n=1}^N\)
        
        \STATE Compute reward $r_k^{n}$ for each $\tilde{\bm{t}}_k^{n}$ 
        and base reward $r^{\text{base}}_k$ for $\bm{t}_k$ using Eq.~\eqref{eq:reward}
        
        \STATE Construct survivors $\mathcal{N}_k^{\text{survive}}$ and calculate $\tilde{\bm{t}}_k$ based on $\bm{t}_k$ and $\mathcal{N}_k^{\text{survive}}$ using Eq.~\eqref{eq:refinement}
        
        \STATE Update global embedding $\bar{\bm{t}}_k$ and counter $n_k$
    \ENDFOR
    
    \STATE Construct prediction embedding set \(\hat{\bm{T}}\) and
    recompute predictions $\{\bm{p}_m\}_{m=1}^M$ using Eq. \eqref{eq:refineprediction}
    
    \STATE Get new pseudo-labels $\{\bar{l}_m\}_{m=1}^M$
    
    \FOR{each high-confidence proposal $\bm{v}_m$}
        \STATE Update $\mathcal{M}_{\bar{l}_m}$ via similarity-based replace
    \ENDFOR
\ENDFOR

\STATE \textbf{Output:} Refined detections $\{\bm{b}_m, \bm{p}_m\}_{m=1}^{M}$ for $x_i$
\end{algorithmic}
\end{algorithm}

\noindent\textbf{Memory update.}
After computing the refined predictions \(\bm{p}_m\), we obtain pseudo-labels by assigning each high-confidence proposal to its predicted class \( \bar{l}_m = \arg\max_k \bm{p}_m[k] \). For each such proposal with visual embedding \(\bm{v}_m\), we update the memory bank \(\mathcal{M}_{\bar{l}_m} = \{ \bm{v}^{{hist}_{\bar{l}_m}} \}_{i=1}^{H_{\bar{l}_m}}\) for its assigned class \(\bar{l}_m\).

Let \(M_{max}\) be the maximum capacity of each class-specific memory. If the current memory size satisfies \(H_{\bar{l}_m} < M_{max}\), we directly append \(\bm{v}_m\) to \(\mathcal{M}_{\bar{l}_m}\). Otherwise, we compute the cosine similarity between \(\bm{v}_m\) and the current global text embedding \(\bar{\bm{t}}_{\bar{l}_m}\), as well as the similarity of all existing memory entries:
\begin{equation}
    \begin{aligned}
s(\bm{v}) = \mathrm{cos}(\bm{v}, \bar{\bm{t}}_{\bar{l}_m}), \quad \bm{v} \in \mathcal{M}_{\bar{l}_m} \cup \{ \bm{v}_m \}.
    \end{aligned}
\end{equation}
Let \(\bm{v}^{\text{min}} = \arg\min_{\bm{v} \in \mathcal{M}_{\bar{l}_m}} s(\bm{v})\) be the least aligned entry. If \(s(\bm{v}_m) > s(\bm{v}^{\text{min}})\), we replace \(\bm{v}^{\text{min}}\) with \(\bm{v}_m\); otherwise, the memory remains unchanged. This ensures that \(\mathcal{M}_k\) always retains at most \(M_{max}\) visual embeddings that are most semantically aligned with the current text embedding.

In summary, RGSE is a lightweight, training-free framework for test-time adaptive object detection under domain shifts, tailored for vision-language models like Grounding DINO (summarized in Algorithm~\ref{alg:method}). Inspired by evolutionary search, it refines text embeddings via four forward-pass stages: perturbation of active classes, reward-based evaluation using current and historical visuals, weighted fusion with global updates, and memory maintenance, ensuring efficient alignment with fallbacks to original embeddings.

\begin{table*}[t]
\centering
\caption{Object Detection Performance Comparison with the state-of-the-art methods on FoggyCityscapes.
Metric: Average Precision@50(\%) - the best numbers are denoted in {\bf bold};
Bp-free: backpropagation-free at test time.}
\label{tab:FoggyCityscapes}
\resizebox{\textwidth}{!}{
\begin{tabular}{l|ccc|c c c c c c c c|c}
\hline
Methods &Venue &Framework &Bp-free&Pson & Rder & Car & Tuck & Bus & Train & Mcle & Bcle & mAP$_{50}$  \\
\hline
\multicolumn{13}{c}{\textbf{Visual Backbone: ResNet-50 }}  \\ \hline
SHOT \cite{liang2020we} &ICML20&Faster RCNN& \XSolidBrush& 26.7 & 30.3 & 36.9 & \bf16.8 & 28.9 & 6.4 & 14.3 & 23.3 & 23.0 \\
T3A \cite{iwasawa2021test} &NIPS21&Faster RCNN& \XSolidBrush& 22.6 & 23.0 & 31.9 & 7.7 & 14.8 & 1.0 & 7.9 & 19.7 & 16.6 \\
Self-Training \cite{xu2021end} &CVPR21&Faster RCNN& \XSolidBrush& 27.7 & 30.8 & 41.4 & 12.8 & 27.4 & 4.2 & 14.8 & 26.1 & 23.1 \\
TTAC \cite{su2022revisiting} &NIPS22&Faster RCNN& \XSolidBrush& 24.5 & 27.3 & 33.4 & 14.6 & 26.1 & 5.8 & 14.1 & 21.5 & 20.9 \\
STFAR\cite{chen2023stfar}  &ARXIV23&Faster RCNN& \XSolidBrush& \bf28.8 & \bf32.0 & \bf42.4 & 15.1 & \bf30.1 & \bf11.2 & \bf15.5 & \bf26.2 & \bf25.1 \\
\hline
\multicolumn{13}{c}{\textbf{Visual Backbone: Swin-T}}  \\ \hline
GDINO \cite{liu2024grounding}&ECCV24&Grounding DINO& \CheckmarkBold &30.10&3.42&46.26&22.41&31.98&0.08&27.96&28.87&23.88 \\
TDA \cite{karmanov2024efficient} &CVPR24&Grounding DINO& \CheckmarkBold&35.27&4.87&46.43&22.82&32.30&0.38&28.68&29.86&25.08\\
HisTPT \cite{zhang2024historical} &NIPS24 &Grounding DINO & \XSolidBrush&32.50&5.09&47.45&23.13&34.63&0.85&28.82&31.98&25.55\\
{BCA} \cite{zhou2025bayesian} &CVPR25&Grounding DINO& \CheckmarkBold & 41.01&5.61&47.85&21.86&33.40&\bf0.91&28.47&29.82&26.12\\
BCA+ \cite{zhou2025bayesianarxiv}&ARXIV25&Grounding DINO& \CheckmarkBold&39.03&3.61&46.98&22.26&35.26&0.87&29.37&35.87&26.65  \\
\hline
RGSE & Ours &Grounding DINO& \CheckmarkBold&\bf41.42&\bf5.70&\bf50.96&\bf28.99&\bf40.50&0.88&\bf32.10&\bf39.10&\bf29.96 \\
\hline
\multicolumn{13}{c}{\textbf{Visual Backbone: Swin-B}}  \\ \hline
GDINO \cite{liu2024grounding} &ECCV24&Grounding DINO& \CheckmarkBold&34.95&20.35&51.56&30.01&45.72&0.87&35.21&32.06&31.34  \\
{TDA} \cite{karmanov2024efficient} &CVPR24&Grounding DINO& \CheckmarkBold 
&38.29&25.53&50.09&30.49&45.92&11.98&33.12&39.06&34.31
\\
HisTPT \cite{zhang2024historical} &NIPS24 &Grounding DINO & \XSolidBrush &37.95&22.46&54.00&29.98&44.11&12.50&37.27&42.33&35.07\\
BCA \cite{zhou2025bayesian} &CVPR25&GroundingDINO& \CheckmarkBold&36.53&24.42&50.77&30.83&45.25&20.29&33.20&42.49&35.47
\\
BCA+ \cite{zhou2025bayesianarxiv}&ARXIV25&Grounding DINO& \CheckmarkBold
&38.15&26.30&52.30&30.77&45.66&20.29&33.89&42.41&36.22
\\ \hline
RGSE & Ours &Grounding DINO& \CheckmarkBold&\bf41.52&\bf27.22&\bf56.78&\bf34.03&\bf50.35&\bf21.68&\bf38.13&\bf47.34&\bf39.63\\\hline
\end{tabular}}
\vspace{-0.4cm}
\end{table*}

\begin{table*}[t]
\centering
\caption{Object Detection Performance Comparison with the state-of-the-art methods on PASCAL-C.
Metric: mean Average Precision@50(\%) -  the best numbers are denoted in {\bf bold}.}
\label{tab:pascalc}
\resizebox{\textwidth}{!}{
\begin{tabular}{l|c c c c c c c c c c c c c c c|c}
\hline
{Methods} & Brit & Contr & Defoc & Elast & Fog & Frost & Gauss & Glass & Impul & Jpeg & Motn & Pixel & Shot & Snow & Zoom & Average \\
\hline
\multicolumn{17}{c}{\textbf{Visual Backbone: ResNet-50 }}  \\ \hline
SHOT \cite{liang2020we} & 72.0 & 31.7 & 18.9 & 46.6 & 67.5 & 45.8 & 12.0 & 11.6 & 16.4 & 41.8 & 19.7 & 33.1 & 19.9 & 42.5 & 27.6 & 33.8 \\
T3A \cite{iwasawa2021test} & 36.9 & 12.5 & 11.0 & 19.7 & 32.7 & 20.6 & 6.1 & 6.4 & 6.5 & 14.8 & 10.1 & 13.2 & 8.4 & 16.8 & 13.8 & 15.3 \\
Self-Training \cite{xu2021end} & 67.9 & 39.3 & 2.6 & 52.5 & 65.7 & 47.2 & 11.9 & 20.2 & 12.1 & 29.3 & 4.1 & 6.9 & 17.4 & 44.9 & 9.5 & 28.8 \\
TTAC \cite{su2022revisiting} & \bf72.2 & 40.4 & 29.3 & \bf58.1 & \bf68.7 & 50.4 & 29.8 & 28.7 & 33.6 & 46.4 & 29.2 & 46.1 & 35.1 & 48.0 & \bf34.9 & 43.4 \\
STFAR \cite{chen2023stfar} & 67.3 & \bf51.8 & \bf34.8 & 55.7 & 65.2 & \bf50.7 & \bf32.4 & \bf34.6 & \bf36.3 & \bf49.4 & \bf34.6 & \bf55.7 & \bf37.8 & \bf50.9 & 34.8 & \bf46.1 \\
\hline
\multicolumn{17}{c}{\textbf{Visual Backbone: Swin-T}}  \\ \hline
GDINO \cite{liu2024grounding} &63.04&45.66&35.58&31.33&60.92&50.52&25.82&18.84&27.99&39.34&26.99&13.39&30.49&45.89&23.38&35.95  \\
TDA \cite{karmanov2024efficient} &64.61&47.71&37.33&32.41&61.77&52.19&27.05&20.48&29.35&41.45&28.50&14.15&31.85&48.07&24.35&37.42\\
HisTPT \cite{zhang2024historical} &66.27&48.10&38.71&34.32&64.24&53.46&28.68&21.75&30.68&41.56&30.20&15.90&33.47&48.46&25.52&38.76\\
{BCA} \cite{zhou2025bayesian} &68.11&49.42&38.46&34.42&64.27&53.21&28.37&22.24&31.90&42.81&30.39&14.81&33.68&49.37&24.97&39.10 \\
BCA+ \cite{zhou2025bayesianarxiv}
&70.25&50.72&40.79&37.08&67.83&56.38&28.59&22.48&31.74&47.40&30.75&14.87&33.66&51.87&25.65&40.67 \\ \hline

RGSE&\bf74.00&\bf54.13&\bf43.02&\bf41.59&\bf72.20&\bf61.04&\bf30.99&\bf25.05&\bf33.86&\bf50.38&\bf34.09&\bf17.65&\bf36.77&\bf56.56&\bf26.59&\bf43.86

\\\hline

\multicolumn{17}{c}{\textbf{Visual Backbone: Swin-B}}  \\ \hline
GDINO  \cite{liu2024grounding}
&85.16&70.69&56.12&55.56&84.56&73.75&54.94&41.17&57.72&71.44&56.08&66.17&60.54&76.48&36.94&63.15
\\
TDA \cite{karmanov2024efficient} &86.28&74.33&57.98&58.45&85.98&76.41&57.82&43.84&60.80&74.31&57.91&69.45&63.79&78.36&38.84&65.64
\\
HisTPT \cite{zhang2024historical} & 89.33&74.89&60.21&59.53&88.14&78.22&58.32&45.48&61.00&75.93&60.13&69.69&64.69&79.79&40.80&67.07\\
{BCA} \cite{zhou2025bayesian} &88.66&75.73&59.84&61.00&88.02&77.69&60.21&45.98&62.46&76.32&60.55&71.22&64.89&80.67&40.66   &67.59
\\
BCA+ \cite{zhou2025bayesianarxiv}
&89.39&77.85&62.04&62.71&88.87&79.94&61.81&47.67&64.87&78.12&61.81&73.51&67.28&82.42&41.42&69.31
\\\hline

RGSE&\bf91.74&\bf82.10&\bf67.23&\bf68.03&\bf92.93&\bf84.29&\bf67.05&\bf52.28&\bf69.87&\bf81.84&\bf67.25&\bf78.92&\bf72.47&\bf86.69&\bf45.53&\bf73.88

\\\hline
\end{tabular}}
\vspace{-0.4cm}
\end{table*}

\section{Experiments}
\subsection{Experimental Setup}

\noindent\textbf{Datasets.}  
Following the evaluation protocol of BCA+~\cite{zhou2025bayesianarxiv}, we evaluate our method on three widely used test-time adaptation benchmarks for object detection:
1. FoggyCityscapes~\cite{sakaridis2018semantic} is a synthetic foggy variant of the Cityscapes dataset~\cite{cordts2016cityscapes}, where realistic fog is added to urban driving scenes. We test on the densest fog level ($\beta = 0.02$).
2. PASCAL-C~\cite{michaelis2019benchmarking} applies 15 types of corruptions to the PASCAL VOC 2007 test set~\cite{everingham2015pascal} (4,952 images) at severity level 5.
3. COCO-C~\cite{michaelis2019benchmarking} applies the same 15 corruptions to 5,000 images from the MS-COCO 2017 validation set~\cite{lin2014microsoft}.
These benchmarks cover a broad spectrum of domain shifts, from adverse weather (FoggyCityscapes) to synthetic corruptions (PASCAL-C, COCO-C), making them ideal for evaluating the robustness of test-time adaptive object detectors.

\begin{table*}[t]
\centering
\caption{Object Detection Performance Comparison with the state-of-the-art methods on COCO-C.
Metric: mean Average Precision@50(\%) -  the best numbers are denoted in {\bf bold}.}
\label{tab:cococ}
\resizebox{\textwidth}{!}{
\begin{tabular}{l|c c c c c c c c c c c c c c c|c}
\hline
{Methods} & Brit & Contr & Defoc & Elast & Fog & Frost & Gauss & Glass & Impul & Jpeg & Motn & Pixel & Shot & Snow & Zoom & Average \\\hline
\multicolumn{17}{c}{\textbf{Visual Backbone: ResNet-50}}  \\ \hline
SHOT \cite{liang2020we} & \bf40.9 & 26.6 & 14.7 & 19.7 & \bf41.5 & 26.7 & 11.0 & 7.2 & 12.1 & 16.4 & 11.0 & 9.7 & 13.0 & 22.0 & 6.4 & 18.6 \\
T3A \cite{iwasawa2021test} & 28.8 & 15.9 & 8.3 & 11.3 & 28.9 & 17.2 & 4.6 & 3.1 & 5.2 & 9.0 & 5.8 & 4.1 & 5.8 & 13.8 & 3.5 & 11.0 \\
Self-Training \cite{xu2021end}& 38.1 & 28.4 & 14.7 & 25.5 & 38.5 & 27.9 & 16.7 & 11.4 & 18.8 & 23.8 & 16.0 & 24.5 & 18.6 & 27.6 & 7.8 & 22.6 \\
TTAC \cite{su2022revisiting} & 38.3 & 29.5 & 15.1 & 28.2 & 39.0 & 28.5 & 16.8 & 14.3 & 18.0 & 23.2 & 14.3 & 24.8 & 19.3 & 26.7 & 8.7 & 23.0 \\
STFAR \cite{chen2023stfar} & 39.1 & \bf31.1 & \bf16.8 & \bf29.0 & 39.0 & \bf29.2 & \bf19.2 & \bf15.4 & \bf20.1 & \bf26.1 & \bf17.2 & \bf28.3 & \bf21.0 & \bf29.5 & \bf10.2 & \bf24.7 \\
W3TTAOD \cite{yoo2024and} & 36.4 & 27.2 & 14.0 & 27.2 & 37.4 & 27.2 & 13.6 & 13.6 & 16.1 & 22.3 & 14.2 & 22.2 & 16.6 & 23.7 & 8.3 & 21.3 \\
\hline

\multicolumn{17}{c}{\textbf{Visual Backbone: Swin-T}}  \\ \hline
GDINO \cite{liu2024grounding} &42.91&23.83&19.45&23.76&43.26&30.36&15.03&10.22&15.87&23.77&16.56&7.31&16.99&25.51&9.35& 21.61 \\
TDA \cite{karmanov2024efficient} &45.53 & 24.43 & 20.33 & 27.08 & 45.91 & 32.31 & 15.77 & 9.93 & 16.72 & 26.25 & 16.85 & 8.99 & 17.10 & 28.22 & 9.62&23.00\\
HisTPT \cite{zhang2024historical} &45.55&26.07&21.25&26.02&45.09&32.82&17.55&12.45&18.21&26.16&19.12&8.62&17.88&28.19&9.93&23.66\\
{BCA} \cite{zhou2025bayesian} &46.34 & 25.18 & 21.03 & 27.45 & 46.74 & 33.92 & 15.97 & 10.91 & 17.33 & 26.19 & 16.05 & 8.89 & 17.97 & 29.02 & 9.87& 23.52  \\
BCA+ \cite{zhou2025bayesianarxiv}
&49.62&26.56&22.20&28.65&49.79&35.65&17.97&11.47&18.67&28.38&19.10&7.83&19.86&30.10&10.08&25.06\\
MPMT \cite{gaotest} &44.9&\bf30.6&17.8&29.9&45.1&34.7&20.2&14.5&21.4&29.2&16.9&\bf23.6&22.0&31.1&7.9&26.0 \\
\hline

RGSE &\bf50.97&29.02&\bf23.76&\bf31.28&\bf53.60&\bf38.91&\bf20.22&\bf15.08&\bf22.12&\bf30.32&\bf20.29&10.79&\bf22.83&\bf33.24&\bf11.25&\bf27.58

\\\hline
\multicolumn{17}{c}{\textbf{Visual Backbone: Swin-B}}  \\ \hline
GDINO \cite{liu2024grounding} 
&55.36&40.40&29.11&36.51&56.26&44.39&30.27&21.38&31.30&40.26&28.78&36.68&32.67&42.71&13.44&35.97
\\
TDA \cite{karmanov2024efficient} 
&56.86&42.26&30.35&38.15&58.21&46.72&32.28&22.64&32.84&42.12&29.94&38.97&34.17&44.14&14.89&37.64
\\
HisTPT \cite{zhang2024historical} &58.02&43.28&31.37&38.51&59.09&46.96&32.85&23.53&33.92&42.23&31.28&39.45&35.06&44.72&15.57&38.39\\
{BCA} \cite{zhou2025bayesian} 
&57.21&42.32&31.00&38.90&58.84&46.43&32.17&23.37&33.53&42.49&31.15&39.56&35.76&45.42&14.92&38.20\\
BCA+ \cite{zhou2025bayesianarxiv}
&60.34&44.99&32.25&40.85&60.92&49.06&34.04&24.42&35.33&44.91&31.72&41.17&36.59&47.85&15.24&39.98
\\

\hline

RGSE&\bf63.81&\bf49.25&\bf34.67&\bf44.64&\bf65.63&\bf52.99&\bf37.22&\bf27.44&\bf38.27&\bf47.86&\bf34.09&\bf44.86&\bf40.28&\bf51.33&\bf16.46&\bf43.25

\\\hline
\end{tabular}}
\vspace{-0.3cm}
\end{table*}

\noindent\textbf{Implementation details.} All experiments are conducted on the PyTorch platform with a batch size of 1, following the standard test-time adaptation protocol. We adopt the Grounding DINO~\cite{liu2024grounding} with Swin-T and Swin-B~\cite{liu2021swin} as the visual backbones, and BERT~\cite{devlin2019bert} as the text encoder. Following BCA+~\cite{zhou2025bayesianarxiv}, we use the hand-crafted prompt template ``\texttt{[class 1] . $\cdots$ . [class C] .}'' for input text encoding. For hyperparameters, we set the number of perturbations to $N = 1000$, the activation threshold to $\tau_{{base}} = 0.7$, the historical weight in the reward to $\alpha = 0.2$, the memory capacity per class to $M_{max} = 2$, and the Gaussian noise standard deviation to $\sigma = 0.1$. All hyperparameters are fixed across datasets. RGSE requires no backpropagation, and the entire adaptation pipeline is executed in a single forward pass per image.

\subsection{Comparisons with State-of-the-Art}

We compare RGSE with state-of-the-art test-time adaptive object detection methods under domain shifts. Traditional methods based on closed-vocabulary detectors like Faster R-CNN with ResNet-50 backbones include SHOT \cite{liang2020we}, T3A \cite{iwasawa2021test}, Self-Training \cite{xu2021end}, TTAC \cite{su2022revisiting}, STFAR \cite{chen2023stfar}, and W3TTAOD \cite{yoo2024and} (on COCO-C only).
In contrast, recent methods leveraging vision-language models (VLMs) like Grounding DINO include GDINO \cite{liu2024grounding}, TDA \cite{karmanov2024efficient}, HisTPT \cite{zhang2024historical}, BCA \cite{zhou2025bayesian}, BCA+ \cite{zhou2025bayesianarxiv} and MPMT \cite{gaotest}.

From Tables~\ref{tab:FoggyCityscapes}--\ref{tab:cococ}, RGSE consistently achieves state-of-the-art mAP$_{50}$ across all three benchmarks under both Swin-T and Swin-B backbones. Specifically, on FoggyCityscapes, RGSE attains 29.96 mAP$_{50}$ with Swin-T and 39.63 with Swin-B, surpassing the strongest prior method (BCA+) by +3.31 and +3.41 points, respectively. On PASCAL-C, RGSE achieves average AP of 43.86 (Swin-T) and 73.88 (Swin-B), outperforming BCA+ by +3.19 and +4.57 points. Similarly, on COCO-C, RGSE yields 27.58 (Swin-T) and 43.25 (Swin-B), improving upon BCA+ by +2.52 and +3.27 points. These gains are also pronounced in challenging categories and corruption types: for instance, RGSE boosts “bicycle” detection by +3.23 (39.10 vs. 35.87) on FoggyCityscapes (Swin-T) and +4.91 (47.34 vs. 42.43) with Swin-B; under “snow” corruption, it improves PASCAL-C by +4.69 (56.56 vs. 51.87, Swin-T) and COCO-C by +3.14 (33.24 vs. 30.10, Swin-T); and for “defocus blur” on PASCAL-C (Swin-T), it achieves 43.02 vs. BCA+’s 40.79, a gain of +2.23. 

Overall, the results across the three tables reveal several key insights: (1) VLM-based methods generally outperform traditional Faster R-CNN approaches, benefiting from linguistic priors that enable better generalization to open-vocabulary and corrupted scenarios, as seen in average AP gaps on PASCAL-C and COCO-C; (2) Among VLM methods, backpropagation-free techniques (e.g., TDA, BCA, BCA+, and RGSE) perform competitively or superiorly to gradient-based ones like HisTPT, underscoring the reliability and efficiency of training-free adaptation, especially on resource-constrained test-time setups; and (3) RGSE's top performance across diverse baselines and backbones validates its effectiveness, with consistent gains in both mean metrics and per-category robustness, proving its value for test time adaptation in vision-language detectors.

\subsection{Ablation Studies}\label{sec:ablation}

\begin{table}[h]
\centering
\caption{Efficiency and performance comparison on COCO-C-Brit. Backbone: Swin-T. 
GPU: NVIDIA A100. Time: average inference time (ms) for 5,000 images. 
Mem: GPU memory (MB).}
\label{tab:efficiency}
\resizebox{0.4\textwidth}{!}{
\begin{tabular}{l|ccc}
\hline
Method  & mAP$_{50}$(\%) & Time(ms) & Mem(MB) \\
\hline
GDINO~\cite{liu2024grounding}      & 42.91 & 147.6 & 3025  \\
TDA~\cite{karmanov2024efficient}   & 45.53 & 411.6 & 4586  \\
HisTPT~\cite{zhang2024historical}  & 45.55 & 1240.8 &12850 \\
BCA+~\cite{zhou2025bayesianarxiv}  & 49.62 & 190.8   & 3862  \\ \hline
{RGSE ($N$=100)} & 48.46& 158.4 & 3152   \\
{RGSE ($N$=500)} & 50.11& 172.8 & 3439\\
{RGSE ($N$=1000)} & 50.97 &182.4 & 3823\\
{RGSE ($N$=1500)} & 51.12 &195.6 & 4190\\
\hline
\end{tabular}}
\vspace{-0.6cm}
\end{table}

\begin{figure*}[h]
    \centering
    \includegraphics[width=\textwidth]{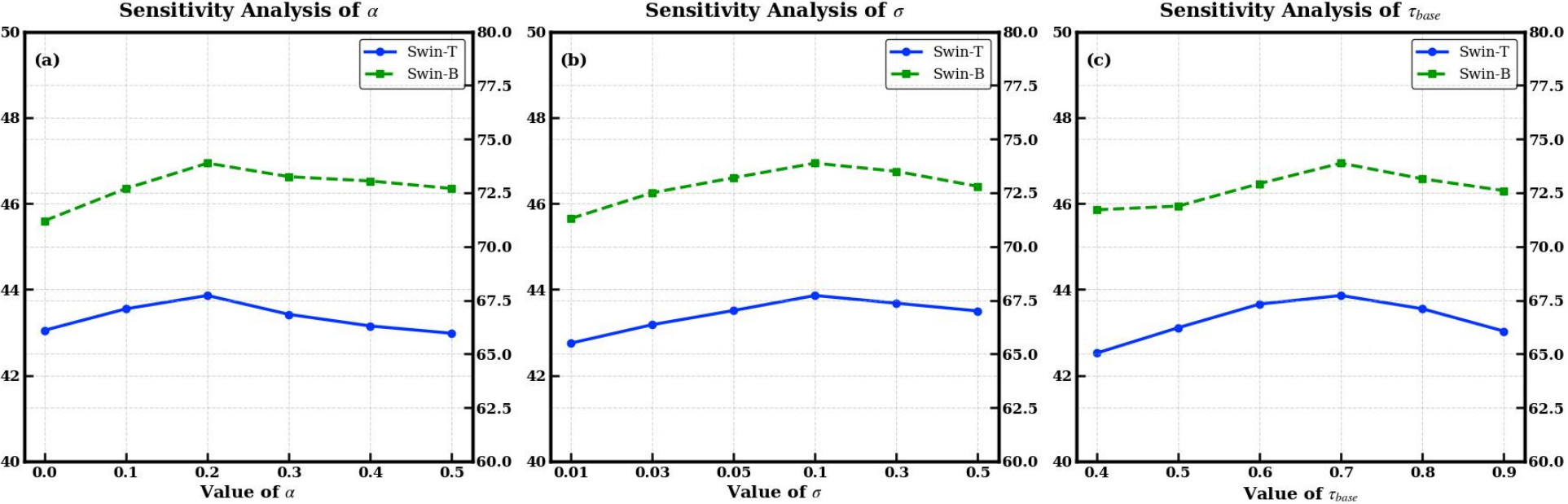}
    \caption{Hyperparameter sensitivity on PASCAL-C. RGSE shows stable performance across a broad range of \(\tau_{\text{base}}\), \(\alpha\), and \(\sigma\).}
    \label{fig:sensitivity}
    \vspace{-0.45cm}
\end{figure*}
\noindent\textbf{Efficiency and performance comparison.} To evaluate the efficiency of RGSE, we compare its inference time and GPU memory usage against state-of-the-art baselines on the COCO-C-Brit benchmark using the Swin-T backbone, measured on an NVIDIA A100 GPU over 5,000 images. As shown in Table~\ref{tab:efficiency}, GDINO serves as a lightweight baseline with 42.91 mAP$_{50}$, requiring 147.6 ms and 3025 MB. TDA improves performance to 45.53 mAP$_{50}$ but increases time to 411.6 ms and memory to 4586 MB. HisTPT achieves a similar mAP$_{50}$ (45.55) but incurs significant overhead, with 1240.8 ms and 12850 MB due to its backpropagation-based updates. BCA+ offers a strong balance at 49.62 mAP$_{50}$, 190.8 ms, and 3862 MB. In contrast, RGSE ($N$=1000) attains mAP$_{50}$ of 50.97 while maintaining competitive efficiency (182.4 ms, 3823 MB), outperforming BCA+ by 1.35 points with less time and memory. This demonstrates RGSE's ability to deliver superior performance without excessive computational costs, making it suitable for real-time test-time adaptation scenarios.

\noindent\textbf{Ablation study on perturbation number $N$.} We conduct an ablation on the number of perturbations $N$ in RGSE to assess its impact on performance and efficiency, as reported in Table~\ref{tab:efficiency}. With $N=100$, RGSE achieves 48.46 mAP$_{50}$ at a low cost (158.4 ms, 3152 MB), already surpassing GDINO and TDA. Increasing to $N=500$ boosts mAP$_{50}$ to 50.11 (+1.65 points) with modest overhead (172.8 ms, 3439 MB). At $N=1000$, mAP$_{50}$ reaches 50.97 (+0.86 points from $N=500$), with time and memory at 182.4 ms and 3823 MB—outperforming BCA+ while remaining efficient. Further raising to $N=1500$ yields diminishing returns (51.12 mAP$_{50}$, +0.15 points) but higher costs (195.6 ms, 4190 MB). Thus, $N=1000$ strikes an optimal balance, expanding the search space effectively without prohibitive overhead, confirming the value of perturbations in RGSE.

\begin{table}[h]
\centering
\caption{Ablation study on memory capacity $M_{max}$ on COCO-C-Brit.}
\label{tab:memory_ablation}
\resizebox{0.4\textwidth}{!}{
\begin{tabular}{l|ccccc}
\hline
$M_{max}$ & 0 & 1 & 2 & 3 & 4\\
\hline
mAP$_{50}$ (\%) & 49.24 & 50.09 & 50.97 & 51.08 & 51.15\\
Mem (MB) &3711&3767&3823& 3879 &3935 \\ \hline
\end{tabular}
}
\end{table}

\noindent\textbf{Ablation study on memory capacity $M_{max}$.}
The memory capacity $M_{\text{max}}$, which controls the maximum number of historical visual embeddings stored per class, is a crucial hyperparameter in RGSE. As shown in Table~\ref{tab:memory_ablation}, increasing $M_{\text{max}}$ from 0 to 4 consistently improves performance, with mAP$_{50}$ rising from 49.24 to 51.15. This demonstrates that incorporating more historical context leads to a more stable and informative reward signal, resulting in better text embedding refinement.
However, the performance gain diminishes significantly when $M_{max} > 2$. The improvement from $M_{max}=2$ to $M_{max}=3$ is only +0.11, and further increasing to $M_{max}=4$ yields a marginal gain of +0.07. In contrast, each increment in $M_{max}$ linearly increases the GPU memory footprint. For every unit increase in $M_{max}$, the system must store an additional visual embedding for all $C$ active categories. While this may be manageable for benchmarks like COCO-C with $C=80$ classes, in real-world open-vocabulary scenarios where the number of potential categories can be much larger (e.g., hundreds or thousands), this overhead can become prohibitively large. Given the rapidly diminishing returns beyond $M_{max}=2$ and the memory cost at scale, we set $M_{max}=2$ as it provides an optimal trade-off between performance and efficiency.

\noindent\textbf{Hyperparameter sensitivity analysis.}
To evaluate the robustness of RGSE to hyperparameter choices, we conduct a sensitivity analysis on PASCAL-C. We vary three key parameters: (1) the activation threshold \(\tau_{{base}}\), (2) the historical weight \(\alpha\) in reward computation, and (3) the standard deviation \(\sigma\) of Gaussian noise for perturbation.
As shown in Figure~\ref{fig:sensitivity}, RGSE achieves consistently strong performance across a wide range of values for all three parameters. This demonstrates that our method is robust to hyperparameter selection, and does not require extensive tuning to achieve high accuracy. The stable performance across settings highlights the practicality of RGSE under diverse test-time conditions.

\begin{figure*}[t]
    \centering
    \includegraphics[width=0.8\textwidth]{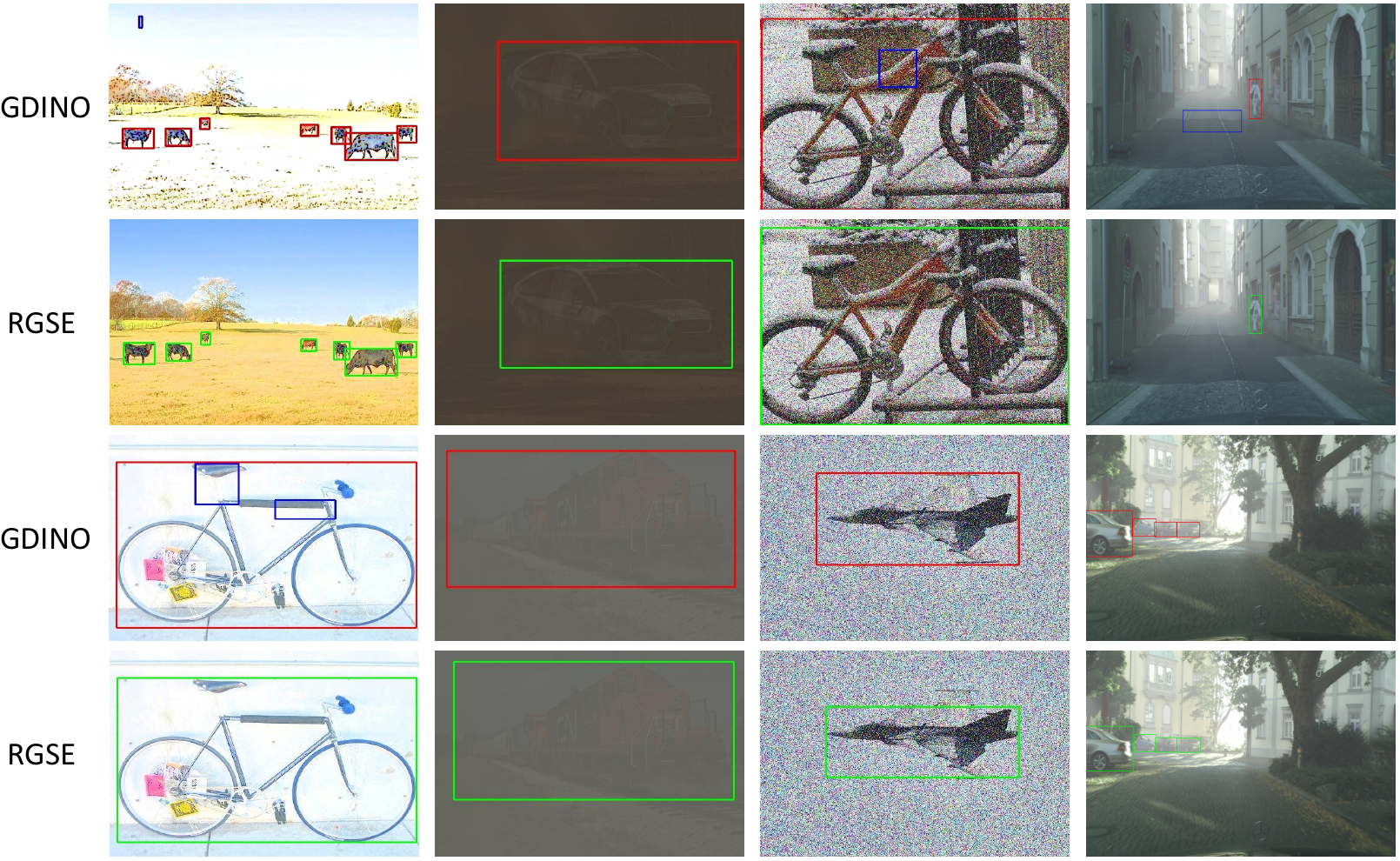}
    \caption{ Qualitative results on PASCAL-C-Brit, PASCAL-C-Contrast, PASCAL-C-GaussNoise, and FoggyCityscapes (Swin-T). Green, red and blue boxes represent true positives, false negatives and false positives, respectively. }
    \label{fig:detections}
    \vspace{-0.4cm}
\end{figure*}

\begin{figure}[t]
  \centering
  \includegraphics[width=0.9\linewidth]{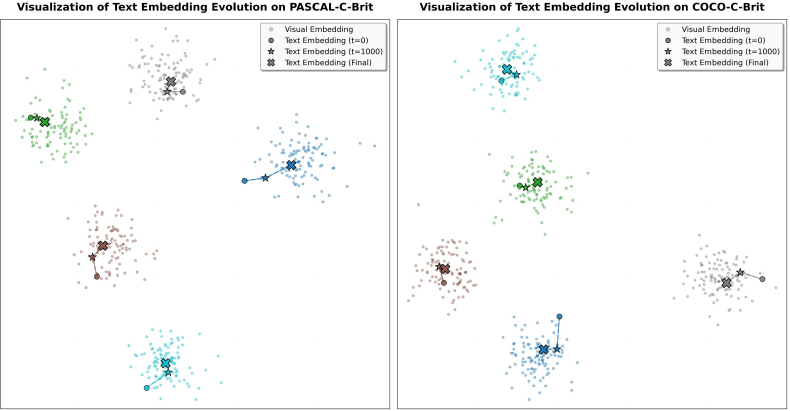} 
  \caption{t-SNE visualization of text embedding trajectories during RGSE adaptation on PASCAL-C-Brit and COCO-C-Brit from $t=0$ to $t=1000$ to final. RGSE pulls text embeddings closer to visual clusters, especially for categories with large initial misalignment.}
  \label{fig:tsne}
  \vspace{-0.4cm}
\end{figure}

\noindent{\bf Qualitative Results on Object Detection.}
To provide a visual comparison, we present qualitative results on the PASCAL-C-Brit, PASCAL-C-Contrast, PASCAL-C-GaussNoise, and FoggyCityscapes datasets using the Grounding DINO model with a Swin-T visual backbone. As shown in Figure~\ref{fig:detections}, we compare the detection outputs of the baseline and our proposed RGSE method.
The visualizations clearly demonstrate that RGSE produces significantly more accurate and robust detections under severe domain shifts. The baseline model frequently fails, suffering from numerous false positives (FP), marked in blue, and missed detections (FN), marked in red. This is particularly evident in challenging conditions such as overexposure (brightness), low contrast, heavy Gaussian noise, and dense fog, where the semantic misalignment between the fixed text embeddings and corrupted visual features leads to poor predictions. In contrast, RGSE achieves superior performance by successfully detecting objects with higher precision and recall. Our method corrects this misalignment through Reward-Guided Semantic Evolution. As a result, RGSE generates the most true positives (TP), marked in green, while effectively suppressing false alarms and recovering missed instances. These qualitative results visually confirm the quantitative superiority of RGSE observed in our experiments.

\noindent{\bf Visualization of Text Embedding Evolution.}
To provide deeper insight into how RGSE progressively corrects semantic misalignment, we visualize the evolution of text embeddings during the adaptation process using t-SNE. This experiment focuses on the trajectory of text embeddings for selected categories under domain shifts, illustrating how RGSE refines them to better align with test-domain visual features. As shown in Figure~\ref{fig:tsne}, we present the t-SNE projections on the PASCAL-C-Brit (categories: aeroplane, bicycle, bird, boat, bottle) and COCO-C-Brit (categories: person, bicycle, car, motorcycle, airplane) datasets using the Grounding DINO model with a Swin-T visual backbone. For each category, we randomly select 100 samples and track the text embeddings at key stages: t=0 (original Grounding DINO features), t=1000 (intermediate after processing 1000 images), and final (after optimization completion on the entire dataset).
The visualizations clearly demonstrate that RGSE enables text embeddings to evolve towards better capturing the visual features in the corrupted domain. At t=0, the original embeddings often show significant separation from visual clusters, reflecting initial misalignment. As adaptation progresses to t=1000 and the final stage, the trajectories converge closer to the visual references, with more pronounced shifts in categories exhibiting large initial differences. This process is particularly evident in challenging brightness corruptions, where RGSE's perturbation and reward-guided refinement adapt the semantic anchors effectively. As a result, the final embeddings achieve tighter alignment, confirming RGSE's ability to enhance semantic coherence and supporting the quantitative cross-modal distance reductions observed in our experiments.

\begin{table}[h]
\centering
\caption{Ablation study on key components of RGSE. Results are averaged over PASCAL-C. Backbone: Swin-T. }
\label{tab:ablation_component}
\resizebox{0.45\textwidth}{!}{
\begin{tabular}{l|ccc|c}
\hline
Method & w/o History & w/o Filtering  & w/o Global Bank & mAP$_{50}$ (\%) \\
\hline
Search-Only Baseline &  & & & 37.95 \\ \hline
 &  &  & \CheckmarkBold & 39.22 \\
 With 1 component&  & \CheckmarkBold &  & 40.06 \\
 & \CheckmarkBold &  &  & 39.91 \\
\hline
 & \CheckmarkBold & \CheckmarkBold &  & 42.89 \\
 With 2 components& \CheckmarkBold &  & \CheckmarkBold & 42.44 \\
 &  & \CheckmarkBold & \CheckmarkBold & 43.05 \\
\hline
RGSE & \CheckmarkBold & \CheckmarkBold & \CheckmarkBold & {43.86} \\
\hline
\end{tabular}
}
\vspace{-0.2cm}
\end{table}

\noindent\textbf{Component analysis.}
To validate the effectiveness of RGSE's core components: Historical Memory (storing visual embeddings from past high-confidence proposals), Filtering (discarding perturbed embeddings with rewards lower than the parent's), and Global Embedding Bank (maintaining a running average of refined embeddings for long-term stability), we conduct an ablation study on PASCAL-C using the Swin-T backbone, averaging results over all corruptions as shown in Table~\ref{tab:ablation_component}. Note that the ``Search-Only Baseline'' is distinct from the zero-shot GDINO; it represents a basic search variant that generates 1,000 perturbed candidates, computes their rewards without historical memory, and fuses them via reward-weighted averaging without survival filtering or global bank updates.
This baseline achieves 37.95 mAP$_{50}$. Adding any single component yields consistent gains: using only the global bank improves performance to 39.22, while reward-based filtering alone gives the largest single-component boost (40.06), and history-aware memory contributes 39.91. When combining any two components, performance further increases demonstrating their complementary roles. The full RGSE model, which integrates all three components, achieves the best result of 43.86 mAP$_{50}$, confirming that history-guided context, selective refinement, and cross-sample embedding accumulation jointly enhance robustness under domain shift.


\section{Conclusion}
In this work, we propose Reward-Guided Semantic Evolution, a training-free framework for test-time adaptive object detection with vision-language models. Unlike existing methods that either rely on backpropagation or bypass text–visual misalignment, RGSE directly refines the text embeddings through evolutionary search: generating perturbed candidates, evaluating them via alignment with current and historical proposals as a reward signal, and fusing only superior variants into a refined embedding.
Extensive experiments show that RGSE achieves state-of-the-art performance across multiple benchmarks while maintaining low computational overhead.

\bibliographystyle{IEEEtran}
\bibliography{cite}

@inproceedings{radford2021learning,
  title={Learning transferable visual models from natural language supervision},
  author={Radford, Alec and Kim, Jong Wook and Hallacy, Chris and Ramesh, Aditya and Goh, Gabriel and Agarwal, Sandhini and Sastry, Girish and Askell, Amanda and Mishkin, Pamela and Clark, Jack and others},
  booktitle={International conference on machine learning},
  pages={8748--8763},
  year={2021},
  organization={PmLR}
}

@inproceedings{liu2024grounding,
  title={Grounding dino: Marrying dino with grounded pre-training for open-set object detection},
  author={Liu, Shilong and Zeng, Zhaoyang and Ren, Tianhe and Li, Feng and Zhang, Hao and Yang, Jie and Jiang, Qing and Li, Chunyuan and Yang, Jianwei and Su, Hang and others},
  booktitle={European Conference on Computer Vision},
  pages={38--55},
  year={2024},
  organization={Springer}
}

@article{pan2009survey,
  title={A survey on transfer learning},
  author={Pan, Sinno Jialin and Yang, Qiang},
  journal={IEEE Transactions on knowledge and data engineering},
  volume={22},
  number={10},
  pages={1345--1359},
  year={2009},
  publisher={IEEE}
}

@article{liang2025comprehensive,
  title={A comprehensive survey on test-time adaptation under distribution shifts},
  author={Liang, Jian and He, Ran and Tan, Tieniu},
  journal={International Journal of Computer Vision},
  volume={133},
  number={1},
  pages={31--64},
  year={2025},
  publisher={Springer}
}

@article{li2025generalizing,
  title={Generalizing Vision-Language Models to Novel Domains: A Comprehensive Survey},
  author={Li, Xinyao and Li, Jingjing and Li, Fengling and Zhu, Lei and Yang, Yang and Shen, Heng Tao},
  journal={arXiv preprint arXiv:2506.18504},
  year={2025}
}

@inproceedings{liang2020we,
  title={Do we really need to access the source data? source hypothesis transfer for unsupervised domain adaptation},
  author={Liang, Jian and Hu, Dapeng and Feng, Jiashi},
  booktitle={International conference on machine learning},
  pages={6028--6039},
  year={2020},
  organization={PMLR}
}

@inproceedings{iwasawa2021test,
  title={Test-time classifier adjustment module for model-agnostic domain generalization},
  author={Iwasawa, Yusuke and Matsuo, Yutaka},
  journal={Advances in Neural Information Processing Systems},
  volume={34},
  pages={2427--2440},
  year={2021}
}

@inproceedings{su2022revisiting,
  title={Revisiting realistic test-time training: Sequential inference and adaptation by anchored clustering},
  author={Su, Yongyi and Xu, Xun and Jia, Kui},
  journal={Advances in Neural Information Processing Systems},
  volume={35},
  pages={17543--17555},
  year={2022}
}

@inproceedings{xu2021end,
  title={End-to-end semi-supervised object detection with soft teacher},
  author={Xu, Mengde and Zhang, Zheng and Hu, Han and Wang, Jianfeng and Wang, Lijuan and Wei, Fangyun and Bai, Xiang and Liu, Zicheng},
  booktitle={Proceedings of the IEEE/CVF international conference on computer vision},
  pages={3060--3069},
  year={2021}
}

@article{chen2023stfar,
  title={Stfar: Improving object detection robustness at test-time by self-training with feature alignment regularization},
  author={Chen, Yijin and Xu, Xun and Su, Yongyi and Jia, Kui},
  journal={arXiv preprint arXiv:2303.17937},
  year={2023}
}

@inproceedings{karmanov2024efficient,
  title={Efficient Test-Time Adaptation of Vision-Language Models},
  author={Karmanov, Adilbek and Guan, Dayan and Lu, Shijian and El Saddik, Abdulmotaleb and Xing, Eric},
  booktitle={Proceedings of the IEEE/CVF Conference on Computer Vision and Pattern Recognition},
  pages={14162--14171},
  year={2024}
}

@inproceedings{zhou2025bayesian,
  title={Bayesian Test-Time Adaptation for Vision-Language Models},
  author={Zhou, Lihua and Ye, Mao and Li, Shuaifeng and Li, Nianxin and Zhu, Xiatian and Deng, Lei and Liu, Hongbin and Lei, Zhen},
  booktitle={Proceedings of the IEEE/CVF conference on computer vision and pattern recognition},
  year={2025}
}

@article{zhou2025bayesianarxiv,
  title={Bayesian Test-time Adaptation for Object Recognition and Detection with Vision-language Models},
  author={Zhou, Lihua and Ye, Mao and Li, Shuaifeng and Li, Nianxin and Wu, Jinlin and Zhu, Xiatian and Deng, Lei and Liu, Hongbin and Luo, Jiebo and Lei, Zhen},
  journal={arXiv preprint arXiv:2510.02750},
  year={2025}
}

@inproceedings{yoo2024and,
  title={What how and when should object detectors update in continually changing test domains?},
  author={Yoo, Jayeon and Lee, Dongkwan and Chung, Inseop and Kim, Donghyun and Kwak, Nojun},
  booktitle={Proceedings of the IEEE/CVF Conference on Computer Vision and Pattern Recognition},
  pages={23354--23363},
  year={2024}
}

@inproceedings{shu2022test,
  title={Test-time prompt tuning for zero-shot generalization in vision-language models},
  author={Shu, Manli and Nie, Weili and Huang, De-An and Yu, Zhiding and Goldstein, Tom and Anandkumar, Anima and Xiao, Chaowei},
  booktitle={Proceedings of the 36th International Conference on Neural Information Processing Systems},
  pages={14274--14289},
  year={2022}
}

@inproceedings{sheng2025r,
  title={R-TPT: Improving Adversarial Robustness of Vision-Language Models through Test-Time Prompt Tuning},
  author={Sheng, Lijun and Liang, Jian and Wang, Zilei and He, Ran},
  booktitle={Proceedings of the Computer Vision and Pattern Recognition Conference},
  pages={29958--29967},
  year={2025}
}

@inproceedings{sharifdeen2025tpt,
  title={O-TPT: Orthogonality Constraints for Calibrating Test-time Prompt Tuning in Vision-Language Models},
  author={Sharifdeen, Ashshak and Munir, Muhammad Akhtar and Baliah, Sanoojan and Khan, Salman and Khan, Muhammad Haris},
  booktitle={Proceedings of the Computer Vision and Pattern Recognition Conference},
  pages={19942--19951},
  year={2025}
}

@inproceedings{feng2023diverse,
  title={Diverse data augmentation with diffusions for effective test-time prompt tuning},
  author={Feng, Chun-Mei and Yu, Kai and Liu, Yong and Khan, Salman and Zuo, Wangmeng},
  booktitle={Proceedings of the IEEE/CVF International Conference on Computer Vision},
  pages={2704--2714},
  year={2023}
}

@article{belal2025vlod,
  title={VLOD-TTA: Test-Time Adaptation of Vision-Language Object Detectors},
  author={Belal, Atif and Medeiros, Heitor R and Pedersoli, Marco and Granger, Eric},
  journal={arXiv preprint arXiv:2510.00458},
  year={2025}
}

@inproceedings{zhang2024historical,
  title={Historical test-time prompt tuning for vision foundation models},
  author={Zhang, Jingyi and Huang, Jiaxing and Zhang, Xiaoqin and Shao, Ling and Lu, Shijian},
  booktitle={Proceedings of the 38th International Conference on Neural Information Processing Systems},
  pages={12872--12896},
  year={2024}
}

@article{hansen2016cma,
  title={The CMA evolution strategy: A tutorial},
  author={Hansen, Nikolaus},
  journal={arXiv preprint arXiv:1604.00772},
  year={2016}
}

@article{jaderberg2017population,
  title={Population based training of neural networks},
  author={Jaderberg, Max and Dalibard, Valentin and Osindero, Simon and Czarnecki, Wojciech M and Donahue, Jeff and Razavi, Ali and Vinyals, Oriol and Green, Tim and Dunning, Iain and Simonyan, Karen and others},
  journal={arXiv preprint arXiv:1711.09846},
  year={2017}
}

@inproceedings{he2016deep,
  title={Deep residual learning for image recognition},
  author={He, Kaiming and Zhang, Xiangyu and Ren, Shaoqing and Sun, Jian},
  booktitle={Proceedings of the IEEE conference on computer vision and pattern recognition},
  pages={770--778},
  year={2016}
}

@article{ren2016faster,
  title={Faster R-CNN: Towards real-time object detection with region proposal networks},
  author={Ren, Shaoqing and He, Kaiming and Girshick, Ross and Sun, Jian},
  journal={IEEE transactions on pattern analysis and machine intelligence},
  volume={39},
  number={6},
  pages={1137--1149},
  year={2016},
  publisher={IEEE}
}

@inproceedings{jia2021scaling,
  title={Scaling up visual and vision-language representation learning with noisy text supervision},
  author={Jia, Chao and Yang, Yinfei and Xia, Ye and Chen, Yi-Ting and Parekh, Zarana and Pham, Hieu and Le, Quoc and Sung, Yun-Hsuan and Li, Zhen and Duerig, Tom},
  booktitle={International conference on machine learning},
  pages={4904--4916},
  year={2021},
  organization={PMLR}
}

@inproceedings{yaofilip,
  title={FILIP: Fine-grained Interactive Language-Image Pre-Training},
  author={Yao, Lewei and Huang, Runhui and Hou, Lu and Lu, Guansong and Niu, Minzhe and Xu, Hang and Liang, Xiaodan and Li, Zhenguo and Jiang, Xin and Xu, Chunjing},
  booktitle={International Conference on Learning Representations}
}

@inproceedings{chenpali,
  title={PaLI: A Jointly-Scaled Multilingual Language-Image Model},
  author={Chen, Xi and Wang, Xiao and Changpinyo, Soravit and Piergiovanni, AJ and Padlewski, Piotr and Salz, Daniel and Goodman, Sebastian and Grycner, Adam and Mustafa, Basil and Beyer, Lucas and others},
  booktitle={The Eleventh International Conference on Learning Representations}
}

@inproceedings{gu2022open,
  title={Open-vocabulary Object Detection via Vision and Language Knowledge Distillation},
  author={Gu, Xiuye and Lin, Tsung-Yi and Kuo, Weicheng and Cui, Yin},
  booktitle={International Conference on Learning Representations},
  year={2022}
}

@inproceedings{yao2022detclip,
  title={DetCLIP: dictionary-enriched visual-concept paralleled pre-training for open-world detection},
  author={Yao, Lewei and Han, Jianhua and Wen, Youpeng and Liang, Xiaodan and Xu, Dan and Zhang, Wei and Li, Zhenguo and Xu, Chunjing and Xu, Hang},
  booktitle={Proceedings of the 36th International Conference on Neural Information Processing Systems},
  pages={9125--9138},
  year={2022}
}

@inproceedings{li2022grounded,
  title={Grounded language-image pre-training},
  author={Li, Liunian Harold and Zhang, Pengchuan and Zhang, Haotian and Yang, Jianwei and Li, Chunyuan and Zhong, Yiwu and Wang, Lijuan and Yuan, Lu and Zhang, Lei and Hwang, Jenq-Neng and others},
  booktitle={Proceedings of the IEEE/CVF conference on computer vision and pattern recognition},
  pages={10965--10975},
  year={2022}
}

@inproceedings{zareian2021open,
  title={Open-vocabulary object detection using captions},
  author={Zareian, Alireza and Rosa, Kevin Dela and Hu, Derek Hao and Chang, Shih-Fu},
  booktitle={Proceedings of the IEEE/CVF conference on computer vision and pattern recognition},
  pages={14393--14402},
  year={2021}
}

@inproceedings{carion2020end,
  title={End-to-end object detection with transformers},
  author={Carion, Nicolas and Massa, Francisco and Synnaeve, Gabriel and Usunier, Nicolas and Kirillov, Alexander and Zagoruyko, Sergey},
  booktitle={European conference on computer vision},
  pages={213--229},
  year={2020},
  organization={Springer}
}

@inproceedings{wang2021tent,
  title={Tent: Fully Test-Time Adaptation by Entropy Minimization},
  author={Wang, Dequan and Shelhamer, Evan and Liu, Shaoteng and Olshausen, Bruno and Darrell, Trevor},
  booktitle={International Conference on Learning Representations}
}

@inproceedings{real2017large,
  title={Large-scale evolution of image classifiers},
  author={Real, Esteban and Moore, Sherry and Selle, Andrew and Saxena, Saurabh and Suematsu, Yutaka Leon and Tan, Jie and Le, Quoc V and Kurakin, Alexey},
  booktitle={International conference on machine learning},
  pages={2902--2911},
  year={2017},
  organization={PMLR}
}

@article{xue2015survey,
  title={A survey on evolutionary computation approaches to feature selection},
  author={Xue, Bing and Zhang, Mengjie and Browne, Will N and Yao, Xin},
  journal={IEEE Transactions on evolutionary computation},
  volume={20},
  number={4},
  pages={606--626},
  year={2015},
  publisher={IEEE}
}

@inproceedings{real2019regularized,
  title={Regularized evolution for image classifier architecture search},
  author={Real, Esteban and Aggarwal, Alok and Huang, Yanping and Le, Quoc V},
  booktitle={Proceedings of the aaai conference on artificial intelligence},
  volume={33},
  number={01},
  pages={4780--4789},
  year={2019}
}

@inproceedings{cheng2024yolo,
  title={Yolo-world: Real-time open-vocabulary object detection},
  author={Cheng, Tianheng and Song, Lin and Ge, Yixiao and Liu, Wenyu and Wang, Xinggang and Shan, Ying},
  booktitle={Proceedings of the IEEE/CVF conference on computer vision and pattern recognition},
  pages={16901--16911},
  year={2024}
}

@inproceedings{liu2021swin,
  title={Swin transformer: Hierarchical vision transformer using shifted windows},
  author={Liu, Ze and Lin, Yutong and Cao, Yue and Hu, Han and Wei, Yixuan and Zhang, Zheng and Lin, Stephen and Guo, Baining},
  booktitle={Proceedings of the IEEE/CVF international conference on computer vision},
  pages={10012--10022},
  year={2021}
}

@inproceedings{devlin2019bert,
  title={Bert: Pre-training of deep bidirectional transformers for language understanding},
  author={Devlin, Jacob and Chang, Ming-Wei and Lee, Kenton and Toutanova, Kristina},
  booktitle={Proceedings of the 2019 conference of the North American chapter of the association for computational linguistics: human language technologies, volume 1 (long and short papers)},
  pages={4171--4186},
  year={2019}
}

@inproceedings{zhangdino,
  title={DINO: DETR with Improved DeNoising Anchor Boxes for End-to-End Object Detection},
  author={Zhang, Hao and Li, Feng and Liu, Shilong and Zhang, Lei and Su, Hang and Zhu, Jun and Ni, Lionel and Shum, Heung-Yeung},
  year={2022},
  booktitle={The Eleventh International Conference on Learning Representations}
}

@article{everingham2015pascal,
  title={The pascal visual object classes challenge: A retrospective},
  author={Everingham, Mark and Eslami, SM Ali and Van Gool, Luc and Williams, Christopher KI and Winn, John and Zisserman, Andrew},
  journal={International journal of computer vision},
  volume={111},
  number={1},
  pages={98--136},
  year={2015},
  publisher={Springer}
}

@article{michaelis2019benchmarking,
  title={Benchmarking robustness in object detection: Autonomous driving when winter is coming},
  author={Michaelis, Claudio and Mitzkus, Benjamin and Geirhos, Robert and Rusak, Evgenia and Bringmann, Oliver and Ecker, Alexander S and Bethge, Matthias and Brendel, Wieland},
  journal={arXiv preprint arXiv:1907.07484},
  year={2019}
}

@article{sakaridis2018semantic,
  title={Semantic foggy scene understanding with synthetic data},
  author={Sakaridis, Christos and Dai, Dengxin and Van Gool, Luc},
  journal={International Journal of Computer Vision},
  volume={126},
  number={9},
  pages={973--992},
  year={2018},
  publisher={Springer}
}

@inproceedings{cordts2016cityscapes,
  title={The cityscapes dataset for semantic urban scene understanding},
  author={Cordts, Marius and Omran, Mohamed and Ramos, Sebastian and Rehfeld, Timo and Enzweiler, Markus and Benenson, Rodrigo and Franke, Uwe and Roth, Stefan and Schiele, Bernt},
  booktitle={Proceedings of the IEEE conference on computer vision and pattern recognition},
  pages={3213--3223},
  year={2016}
}

@inproceedings{lin2014microsoft,
  title={Microsoft coco: Common objects in context},
  author={Lin, Tsung-Yi and Maire, Michael and Belongie, Serge and Hays, James and Perona, Pietro and Ramanan, Deva and Doll{\'a}r, Piotr and Zitnick, C Lawrence},
  booktitle={European conference on computer vision},
  pages={740--755},
  year={2014},
  organization={Springer}
}

@inproceedings{wang2025efficient,
  title={Efficient Test-time Adaptive Object Detection via Sensitivity-Guided Pruning},
  author={Wang, Kunyu and Fu, Xueyang and Lu, Xin and Ge, Chengjie and Cao, Chengzhi and Zhai, Wei and Zha, Zheng-Jun},
  booktitle={Proceedings of the Computer Vision and Pattern Recognition Conference},
  pages={10577--10586},
  year={2025}
}

@inproceedings{zhai2022lit,
  title={Lit: Zero-shot transfer with locked-image text tuning},
  author={Zhai, Xiaohua and Wang, Xiao and Mustafa, Basil and Steiner, Andreas and Keysers, Daniel and Kolesnikov, Alexander and Beyer, Lucas},
  booktitle={Proceedings of the IEEE/CVF conference on computer vision and pattern recognition},
  pages={18123--18133},
  year={2022}
}

@inproceedings{cui2022contrastive,
  title={Contrastive vision-language pre-training with limited resources},
  author={Cui, Quan and Zhou, Boyan and Guo, Yu and Yin, Weidong and Wu, Hao and Yoshie, Osamu and Chen, Yubo},
  booktitle={European Conference on Computer Vision},
  pages={236--253},
  year={2022},
  organization={Springer}
}

@inproceedings{vs2023towards,
  title={Towards online domain adaptive object detection},
  author={VS, Vibashan and Oza, Poojan and Patel, Vishal M},
  booktitle={Proceedings of the IEEE/CVF Winter Conference on Applications of Computer Vision},
  pages={478--488},
  year={2023}
}

@inproceedings{zhang2024dual,
  title={Dual prototype evolving for test-time generalization of vision-language models},
  author={Zhang, Ce and Stepputtis, Simon and Sycara, Katia and Xie, Yaqi},
  journal={Advances in Neural Information Processing Systems},
  volume={37},
  pages={32111--32136},
  year={2024}
}

@inproceedings{gaotest,
  title={Test-Time Adaptive Object Detection with Foundation Model},
  author={Gao, Yingjie and Zhang, Yanan and Cai, Zhi and Huang, Di},
  booktitle={The Thirty-ninth Annual Conference on Neural Information Processing Systems},
  year={2025}
}

@inproceedings{ruan2024fully,
  title={Fully test-time adaptation for object detection},
  author={Ruan, Xiaoqian and Tang, Wei},
  booktitle={Proceedings of the IEEE/CVF Conference on Computer Vision and Pattern Recognition},
  pages={1038--1047},
  year={2024}
}

@article{liu2024mlfa,
  title={MLFA: Toward realistic test time adaptive object detection by multi-level feature alignment},
  author={Liu, Yabo and Wang, Jinghua and Huang, Chao and Wu, Yiling and Xu, Yong and Cao, Xiaochun},
  journal={IEEE Transactions on Image Processing},
  volume={33},
  pages={5837--5848},
  year={2024},
  publisher={IEEE}
}

@article{liu2025towards,
  title={Towards Efficient Test time Adaptation with Hierarchical Distribution Alignment},
  author={Liu, Yabo and Huang, Chao and Xu, Yong and Cao, Xiaochun and Wang, Jinghua},
  journal={IEEE Transactions on Image Processing},
  year={2025},
  publisher={IEEE}
}

@article{zhang2025cola,
  title={COLA: Context-aware Language-driven Test-time Adaptation},
  author={Zhang, Aiming and Yu, Tianyuan and Bai, Liang and Tang, Jun and Guo, Yanming and Ruan, Yirun and Zhou, Yun and Lu, Zhihe},
  journal={IEEE Transactions on Image Processing},
  year={2025},
  publisher={IEEE}
}

@article{wu2025a3,
  title={A3-TTA: Adaptive Anchor Alignment Test-Time Adaptation for Image Segmentation},
  author={Wu, Jianghao and Luo, Xiangde and Zhou, Yubo and Wu, Lianming and Wang, Guotai and Zhang, Shaoting},
  journal={IEEE Transactions on Image Processing},
  volume={34},
  pages={8511--8522},
  year={2025},
  publisher={IEEE}
}

@article{tian2026dual,
  title={Dual Domain-attribute Learning Framework with Asynchronous Adapters for Continual Test-time Adaptation},
  author={Tian, Yuntong and Li, Kang and He, Tianyang and Wan, Liang and Heng, Pheng-Ann and Feng, Wei},
  journal={IEEE Transactions on Image Processing},
  year={2026},
  publisher={IEEE}
}

@article{shao2025consistent,
  title={Consistent Assistant Domains Transformer for Source-Free Domain Adaptation},
  author={Shao, Renrong and Zhang, Wei and Luo, Kangyang and Li, Qin and Wang, Jun},
  journal={IEEE Transactions on Image Processing},
  year={2025},
  publisher={IEEE}
}

@article{wang2025deep,
  title={Deep label propagation with nuclear norm maximization for visual domain adaptation},
  author={Wang, Wei and Li, Hanyang and Wang, Cong and Huang, Chao and Ding, Zhengming and Nie, Feiping and Cao, Xiaochun},
  journal={IEEE Transactions on Image Processing},
  volume={34},
  pages={1246--1258},
  year={2025},
  publisher={IEEE}
}

@article{lu2025adaptive,
  title={Adaptive dispersal and collaborative clustering for few-shot unsupervised domain adaptation},
  author={Lu, Yuwu and Huang, Haoyu and Wong, Wai Keung and Hu, Xue and Lai, Zhihui and Li, Xuelong},
  journal={IEEE Transactions on Image Processing},
  year={2025},
  publisher={IEEE}
}

\end{document}